\documentclass[10pt,twocolumn,letterpaper]{article}

\usepackage{cvpr}              

\usepackage{graphicx}
\usepackage{amsmath}
\usepackage{amssymb}
\usepackage{booktabs}
\usepackage{wrapfig}

\usepackage[pagebackref,breaklinks,colorlinks]{hyperref}

\usepackage[capitalize]{cleveref}
\crefname{section}{Sec.}{Secs.}
\Crefname{section}{Section}{Sections}
\Crefname{table}{Table}{Tables}
\crefname{table}{Tab.}{Tabs.}

\def\cvprPaperID{12045} 
\def\confName{CVPR}
\def\confYear{2023}

\def\eg{\emph{e.g.~}}

\def\ie{\emph{i.e.~}}

\begin{document}

\title{BinaryVQA: A Versatile Test Set to Evaluate the Out-of-Distribution Generalization of VQA Models}

\author{Ali Borji\\
Quintic AI\\
{\tt\small aliborji@gmail.com}
}
\maketitle

\begin{abstract}
We introduce a new test set for visual question answering (VQA) called BinaryVQA to push the limits of VQA models. Our dataset includes 7,800 questions across 1,024 images and covers a wide variety of objects, topics, and concepts. For easy model evaluation, we only consider binary questions. Questions and answers are formulated and verified carefully and manually. Around 63\% of the questions have positive answers. The median number of questions per image and question length are 7 and 5, respectively. The state of the art OFA model achieves 75\% accuracy on BinaryVQA dataset, which is significantly lower than its performance on the VQA v2 test-dev dataset (94.7\%). We also analyze the model behavior along several dimensions including: a) performance over different categories such as text, counting and gaze direction, b) model interpretability, c) the effect of question length on accuracy, d) bias of models towards positive answers and introduction of a new score called the ``ShuffleAcc'', and e) sensitivity to spelling and grammar errors. Our investigation demonstrates the difficulty of our dataset and shows that it can challenge VQA models for next few years. 
Data and code are publicly available at: \href{https://drive.google.com/file/d/18g0AV6xdSGGMc4Bg0vQ2U6KR7K1_Jo6W/view?usp=sharing}{DATA} and  \href{https://github.com/aliborji/BinaryVQA}{CODE}.




\end{abstract}

\vspace{-15pt}
\section{Introduction}

\begin{figure}[t] 
\centering
  \includegraphics[width=.95\linewidth]{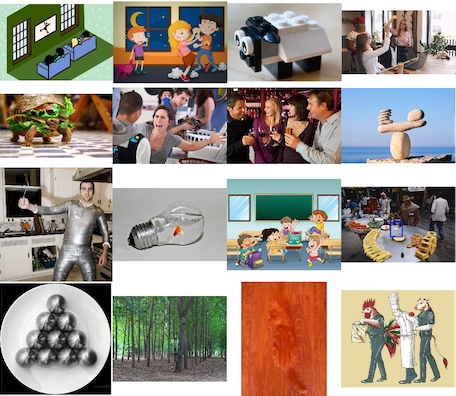} 
\caption{Samples from our dataset. Our dataset covers a wide variety of concepts including counting, crowd, emotions, drawings, paintings, camouflage, clothing, time, weather, body parts, age, text, gaze direction, etc. It also includes questions that address spatial understanding of models (\eg the blue rectangle in the last image of the 3rd row). See Appendix A for more examples.}
\label{fig:samples}
\vspace{-10pt}
\end{figure}

Visual question answering~\cite{antol2015vqa,geman2015visual} is a multidisciplinary task at the intersection of computer vision, NLP, knowledge representation, reasoning, common sense knowledge, etcetra. The goal is to answer a text-based question given an input still image or a video. 

Recent VQA models are able to answer binary questions above 95\% accuracy, which is astonishing considering that in principle, any questions can be asked on an image. At the same time, though, this alarms that perhaps we are not using the test sets that have the right level of difficulty. Using the same test set over the years has the risk of over-fitting, as researchers often tune their models towards the statistics of the test sets (even when the annotations are held hidden). To mitigate this issue, it is crucial to have several versatile independent test sets to evaluate models and to track the progress. While several test sets are available for problems such as image classification (\eg~\cite{hendrycks2021nae,recht2019imagenet,barbu2019objectnet}) and object detection (\eg~\cite{lau2021natural,lin2014microsoft,krasin2017openimages}), the VQA field lacks enough difficult test sets. Our study is an effort in this direction. This discussion naturally relates to the out-of-distribution studies showing that models are biased towards the test sets that are similar to the sets over which they have been trained on. Likewise, they underperform over test sets that are even slightly different~\cite{recht2019imagenet,shankar2020evaluating,taori2020measuring}. In this regard, here we also testing the out-of-distribution performance of the VQA models.


\begin{table*}[t]
\vspace{-10pt}
\centering
\renewcommand{\tabcolsep}{5pt}
\renewcommand{\arraystretch}{.8}
\begin{footnotesize}
\begin{tabular}{|lcc|c|}
\hline
\textbf{Dataset} & \textbf{\# Images} & \textbf{\# Questions} & \textbf{Question Type(s)}  \\
\hline
\hline
DAQUAR \cite{malinowski2014multi}	&1449	&12468	&Object identification	 \\
\hline
COCO-QA \cite{ren2015exploring}	&123287	& 115000	& Questions automatically generated from COCO captions \\
\hline
VQA \cite{antol2015vqa}	&204721	&614163	&Combining vision, language and common-sense \\
\hline
Visual Madlibs \cite{yu2015visual}	&10738	&360001	&Fill in the blanks	\\
\hline
Visual7W \cite{zhu2016visual7w}	&47300	&2201154	& 7Ws, locating objects  \\
\hline
CLEVR~\cite{johnson2017clevr}	&100000	&853554	&Synthetic question generation using relations	 \\
\hline
Tally-QA \cite{acharya2019tallyqa}	&165000	&306907	&Counting objects on varying complexities \\
\hline
KVQA \cite{shah2019kvqa}	&24602	&183007	&Questions based on Knowledge Graphs \\
\hline
VizWiz \cite{gurari2018vizwiz}	& 31000	& 31000	& Questions by visually impaired users \\
\hline
TextVQA \cite{singh2019towards}	& 28408 & 45336	& Questions demanding reasoning about text  \\
\hline
\end{tabular}
\end{footnotesize}
\vspace{-5pt}
\caption{Overview of VQA datasets described in this paper.}
\vspace{-10pt}
\label{tab:datasets}
\end{table*}

Our test set contains 1,024 images crawled from publicly-available and free-to-distribute sources. We used Google and Bing search engines with different search phrases to collect the images. We made sure that no image contains sensitive material, has poor resolution, or violates the copyright law\footnote{We choose images that were public domain, did not have copyright, or were released by the government.}. The gathered data encompass a wide variety of visual concepts over both RGB images, paintings, drawings, cartoons, and clip arts (Fig.~\ref{fig:samples}). We have made sure that all the questions are unambiguous and answers are correct. Our test set contains more questions per image ($\sim$7) than the VQA v2 test set ($\sim$3). We only consider the binary questions, since essentially any question can be converted to a ``yes/no'' question. This simplifies the model evaluation and eliminates
the complicated process of matching sentences of predicted answers with actual answers. Notice that this argument does not necessarily mean that we only need models that give binary answers.

Although our test set is smaller than the VQA test set, it comes with the benefit of better control over the complexity of the questions and quality of the answers. Controlling the difficulty level of the questions generated by the Amazon Mechanical Turk (AMT) workers is challenging, as workers may choose to ask simple and short questions to save time. Unlike the questions in the VQA dataset~\cite{antol2015vqa} that are supposed to fool a toddler, alien, or a smart robot, some BinaryVQA questions can even challenge adults. To answer the majority of the questions, one has to carefully analyze the images. Further, small versatile and carefully curated test sets like ours can alleviate the legal issues concerning consents, licensing, privacy and security which are harder to control in datasets containing millions of images.




In curating the BinaryVQA, we have made three choices. First, this test set is intentionally not paired with a training set. This is to encourage generalization and
to prohibit models to take advantage of correlations between testing and training sets. 
These correlations are easily accessible to models but are not detectable by humans~\cite{geirhos2020shortcut}.
Second, our dataset comes with a license that disallows researchers to update the parameters of any model for any reason on it. This is again to avoid over-fitting. Third, to mitigate the danger of leaking our data to other training sets, we mark every image by a one pixel green border that must be removed on the fly before testing.



In addition to the test set, we also introduce new dimensions along which VQA models can be tested, in particular sensitivity of the models to small perturbations in the questions. We find that, unlike humans, current models are highly sensitive to minor grammar mistakes. Further, we study the bias of models towards generating positive answers, whether models indeed require the image to answer the questions, and whether they choose the right image regions to do so. In a nutshell, our results show that state of the art VQA models struggle on our dataset. This suggests that, in conjunction with other datasets, our dataset can be used to push the VQA models to become better.

\section{VQA Datasets}

Several VQA datasets have been introduced~\cite{wu2017visual,kafle2017visual,manmadhan2020visual}.
In these datasets, images are either taken from an existing vision dataset (\eg MSCOCO;~\cite{lin2014microsoft}) or are artificially created (\eg Abstract Scenes;~\cite{antol2015vqa}, computer graphics;~\cite{andreas2016neural,johnson2017clevr}). Further, questions are generated either automatically~\cite{andreas2016neural,johnson2017clevr,kafle2017visual,malinowski2014multi,ren2015exploring,yu2015visual}, from crowd workers~\cite{antol2015vqa,gao2015you,goyal2017making,kafle2017visual,krishna2017visual,zhu2016visual7w}, or from in-house participants~\cite{kafle2017visual,wang2015explicit}. Unlike these datasets, questions in our dataset are carefully constructed by experts such that to answer them a detailed inspection of the image is necessary. Some prominent VQA datasets are listed in Table~\ref{tab:datasets}. Relevant ones to our work are described next.


{\bf COCO-QA~\cite{ren2015exploring}} includes 123,287 images from the MSCOCO (72,783 for training and 38,948 for testing) and each image has one question/answer pair. Questions are automatically generated from the image descriptions and are categorized into four types based on the type of expected answer: object, number, color, and location. A downside of the COCO-QA dataset is that 9,072 (23.29\%) of test questions also appear in the training questions.






{\bf VQA~\cite{antol2015vqa,goyal2017making}} is one of the most widely used datasets (\url{https://visualqa.org/}). It comprises two parts, one using natural images called VQA-real (sourced from MSCOCO), and a second one with cartoon images called VQA-abstract. The latest more comprehensive version of this dataset, VQA v2.0 consists of 1.1 million (image, question) pairs with 13 million associated answers. 

{\bf Visual Genome~\cite{krishna2017visual}} is aimed to enhance the progress on cognitive
tasks, especially spatial relationship reasoning. It contains over 108K images, with about 35 objects, 26 attributes, and 21 pairwise relationships between objects. 


{\bf Visual7W~\cite{zhu2016visual7w}} includes seven types of WH questions (what, where, when, who, why, which and how) to examine capability of a model in visual understanding. 
Questions are asked in the multiple-choice format. There are four candidates for
each question, and only one candidate is the correct answer.

{\bf Visual Madlibs~\cite{yu2015visual}} consists of 360,001 targeted descriptions spanned across 12 different types of templates and
their corresponding images.


{\bf VizWiz~\cite{gurari2018vizwiz}} is constructed from interactions of visually impaired users with a mobile application. It consists of 31,000 visual questions together with 10 crowdsourced
answers per question. Images often have poor quality due to poor lighting, focus, and framing of the content of interest. Further, questions are on average more conversational and are sometimes incomplete. 

{\bf TextVQA~\cite{singh2019towards}} contains 45,336 questions on 28,408 images that require reasoning about text to be answered. Images are taken from the Open Images v3 dataset~\cite{krasin2017openimages}. TextVQA is available at \url{https://textvqa.org}.





In addition to above, some non-photo-realistic datasets such as CLEVR~\cite{johnson2017clevr},
NLVR~\cite{suhr2017corpus}, and FigureQA~\cite{kahou2017figureqa} have also been introduced to study visual reasoning independent of language. Some datasets such as Fact-Based VQA~\cite{wang2017fvqa} explicitly require external knowledge to answer questions. 
GQA~\cite{hudson2019gqa} is a popular dataset, which also involves phrases to address the relations.

Our work relates to research that addresses the functional diagnostics of pre-trained language models (\eg~\cite{rottger2020hatecheck,nangia2020crows}). It also relates to works that examine adversarial robustness and out-of-distribution generalization of VQA models (\eg~\cite{li2021adversarial,bugliarello2021multimodal}). For example, \cite{li2021adversarial} shows that non-expert annotators can easily attack the best VQA models.


We construct an adversarial dataset to challenge the best VQA models. Although there are few such datasets for free-form VQA (\eg VQA-CP~\cite{agrawal2018don}), here we show that even that answering yes/no questions is not yet solved. 

\section{BinaryVQA Dataset}
\begin{figure*}[t]
\centering
  \hspace{-10pt}
  \vspace{-15pt}
  
  \includegraphics[width=.55\linewidth]{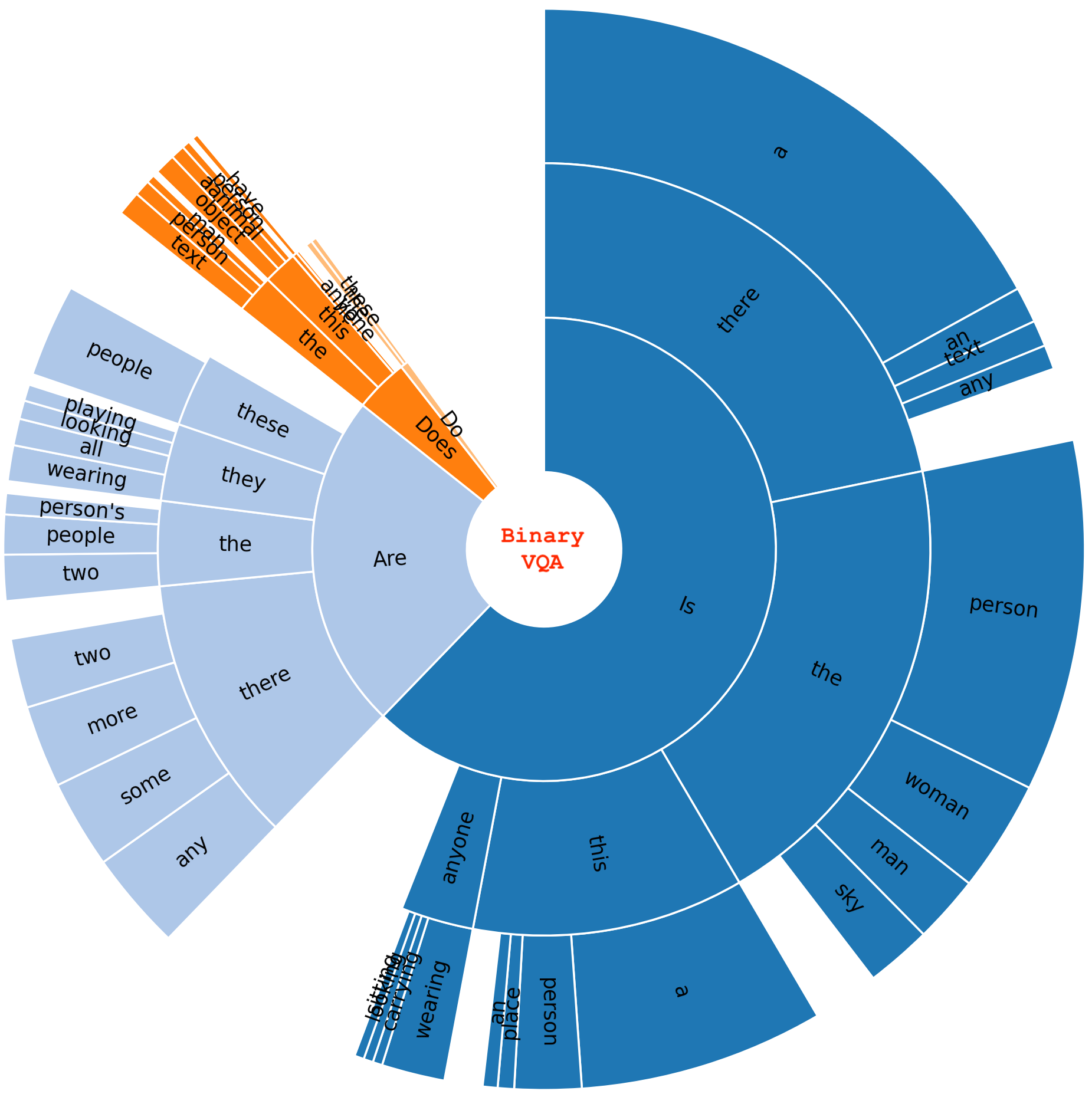} \hspace{15pt}
  \includegraphics[width=.35\linewidth]{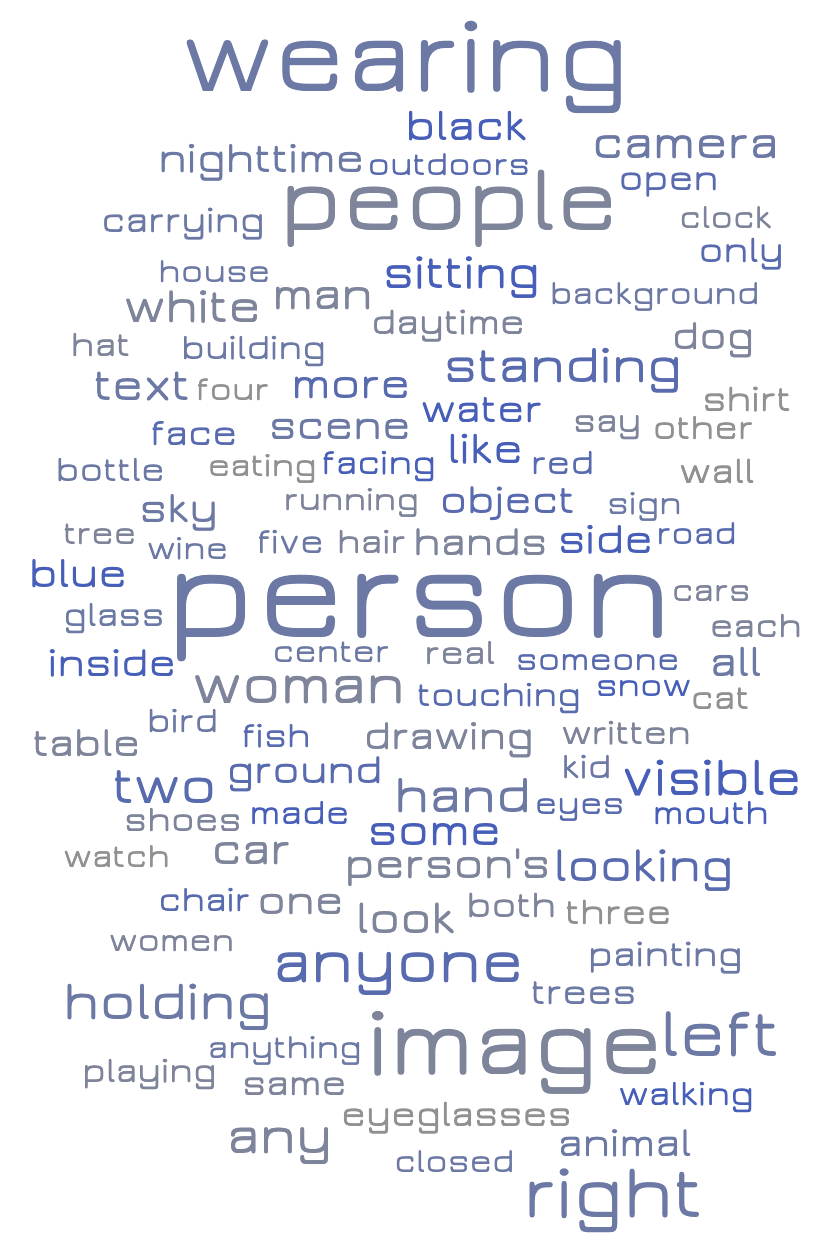}   
\caption{Left: Distribution of questions in our dataset by their first three words. The ordering of the words starts towards the center and radiates outwards. The arc length is proportional to the number of questions containing the word. Right: Venn-style word clouds of words in the questions. The most frequent word is `person' indicating that questions are often about people in the images.}
\label{fig:sunburst}
\vspace{-5pt}
\end{figure*}

Our dataset contains 7,800 questions across 1,024 images. Majority of the questions start with ``Is'' and ``Are'' as shown in the sunburst plot in Fig.~\ref{fig:sunburst}. The most common terms in the questions are \texttt{person, wearing, people, and image} (right panel in Fig.~\ref{fig:sunburst}). We do not include WH questions and all questions have ``yes'' or ``no'' answers.
We ensured that each image is valid through human review. We formulated the questions and then presented them along with their answers to three AMT workers for verification. Please see Appendix D for details. Out of all questions, only 41 QA pairs received the incorrect majority vote, which were fixed subsequently.

\begin{figure*}
\centering
  \hspace{-5pt}
  \includegraphics[width=.33\linewidth]{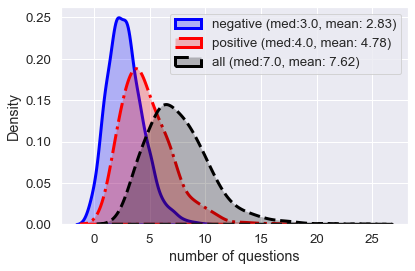} 
  \includegraphics[width=.33\linewidth]{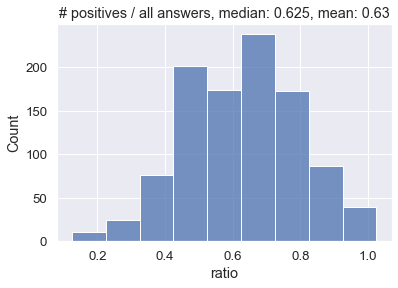} 
  \includegraphics[width=.33\linewidth]{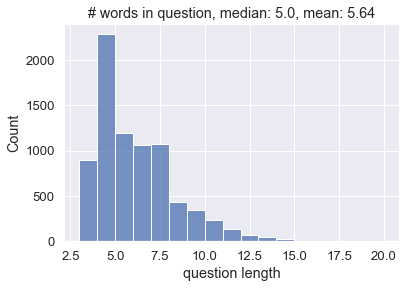} 
\vspace{-7pt}
\caption{BinaryVQA dataset statistics. Left: Distribution of the number of questions and its breakdown on positive and negative answers. Half of the images have more than 7 questions. Middle: Ratio of positive to all questions. On average images contain more positive questions than negative ones. Right: Distribution of question length. Half of the questions have length greater than five.}
\label{fig:stats}
\vspace{-10pt}
\end{figure*}

\begin{table}
\begin{center}
\begin{scriptsize}
\renewcommand{\tabcolsep}{2pt}
\renewcommand{\arraystretch}{.5}
\begin{tabular}{|l|l|}
\hline 
{\bf Q type} & {\bf List of words} \\
\hline \hline 
sky & sky\\
spatial & rectangle\\
vegetation & tree, plant, flower\\
gaze direction & looking\\
real/drawing & painting, drawing\\
$[$in/out$]$doors & indoors, outdoors\\
daytime & daytime, nighttime\\
emotions & happy, sad, angry, upset\\
time & clock, time, watch, hour, minute, seconds\\
gender & man, woman, female, male, boy, girl\\
text & text, number, English, Roman, word, written\\
age & age, old, young, child, kid, baby, adult, teenager\\
weather & weather, snowy, sunny, cloudy, rainy, stormy, foggy\\
color & color, white, red, blue, yellow, black, purple, green, silver, blond\\
actions & fighting, walking, sitting, standing, running, climbing, lying,\\
& dancing, partying\\
direction & right, left, top, bottom, above, below, side, leftmost, rightmost, next\\
counting & more than, less than, two, three, ten, fifteen, twenty, \\
& two hundred, exactly, only\\
body parts & face, head, hand, leg, foot, feet, eye, torso, ear, \\
& belly, belly button, finger, hair, shoulder, neck, mouth, nose, body\\
clothing & shoe, jean, jeans, dress, tie, shirt, short, long sleeve, sock, hat, cap, \\
& earring, watch, piercing, necklace, scarf, eyeglasses,\\
& belt, cloths, wearing\\
animals & animal, cat, dog, elephant, tiger, horse, owl, chicken, hen, \\
& rooster, wolf, fox, octopus, sheep, eagle, lion, giraffe, monkey, \\
& cow, scorpion, turtle, fly, mosquito, dinosaur, panda, pigeon, spider\\
fruits & fruit, apple, banana, acorn, tomato, potato, pomegranate, \\
& pear, peach, orange, grape, melon, watermelon, cherry, strawberry, \\
& corn, pumpkin, pineapple, lemon,pepper, avocado, cabbage, \\
& lettuce, coconut, cucumber, eggplant, broccoli\\
\hline 
\end{tabular}
\end{scriptsize}
\end{center}
\vspace{-10pt}
\caption{List of words per question type in the BinaryVQA dataset.}
\vspace{-15pt}
\label{tab:qtypes}
\end{table}



%

Statistics of the BinaryVQA dataset are shown in Fig.~\ref{fig:stats}. Out of the 7,800 questions, 4,897 have positive answers and the remaining 2,903 have negative answers, resulting in a ratio of about 62.7\% (positive/all images). The median positive to all questions ratio per image is 0.625. 38 images (3.7\%) have all of their questions answered ``yes'', while no image has all of its questions answered ``no''. The median number of questions per image is 7 which means that half of the images have more than 7 questions.
The median number of positive questions (questions with answer ``yes'') is 4 and the median number of negative questions is 3. The mean number of questions per image in BinaryVQA is 7.62 which is higher than 5.4 for VQA v2. BinaryVQA questions range from 3 to 20 words. The mean and median question length are 5.64 and 5 words, respectively. VQA v2 questions range from 4 to 10 words (average 5). The average image resolution is 840.3 $\times$ 650.4 (w $\times$ h) with the average aspect ratio of 1.32.


Sample images are shown in Fig.~\ref{fig:samples}. BinaryVQA images and questions cover a wide variety of topics and concepts including drawings, paintings, uncommon views of objects, hybrid animals, out of context objects and odd scenes (elephant in the room, car in the swimming pool, black sheep among white sheep), weather conditions, time, interactions among people, actions (fighting, running, walking, dancing), emotions (sadness, happiness, surprise, anger), counts and quantity, gender, age, race, gaze direction, object materials, objects in the mirror, body parts (\eg whether mouth or eyes are open, whether teeth are visible), animals, fruits, clothing (T-shirt, long sleeve, pants), shadow, color, crowd, clouds, tattoos, camouflage, illusions, non-existing objects, and logical reasoning.

In formulating the questions, we tried to remove any ambiguity (\eg in giving addresses relative to the image, objects, people in the scene, or image viewer; left side of the rightmost person; left of the image). When only some people in the image (\eg standing ones) are doing an action, we did not ask ``Are these people doing X''. Instead, we asked ``Are the standing people in this image doing X''. 


Some questions test whether models can tell the type of the image (\eg ``Is this a drawing?'' and ``Is this a painting?'') and whether they can answer questions over different types of images (\eg drawings, paintings, cartoons, clip art, black and white images). Some questions ask about the text, for example ``Is there text?'', ``Is the word X written somewhere in this image?'', ``Is the text written in English?'', ``Is the number 53813 written somewhere in the image?''. External knowledge and common sense are needed to answer some questions (\eg ''Is this a map of Japan?, ``Is this person a celebrity?''). In order to further test the spatial understanding of the models, we placed a blue rectangle around some objects in the image and targeted the questions only on those regions (See Fig.~\ref{fig:samples}). An example question is ``Is the spatula inside the blue rectangle blue?''. To test the consistency of models and see whether they truly understand the image, for some images we include questions that contradict each other (\eg ``Is the boy standing?'' \textit{vs} ``Is the boy sitting?''). Some other sample questions are ``Is the whole body of the person visible?'', ``Is she holding a wine in her left hand?'', ``Are some birds printed on her skirt?'', ``Is her right hand in her right pocket?'', ``Is the person on the left taller?'', ``Is anyone looking at the camera?'', Is this person an adult?'', ``Is the sky clear?'', ``Are his feet touching the ground?'', ``Are there more X objects than Y objects?'', ``Is object X to the left of object Y?'', ``Is the person in the image female?'', and ``Is the person opening the door with his right hand?''. We clustered the questions based on the terms that appeared in them, as shown in Table~\ref{tab:qtypes}. For example, questions with words \texttt{gender, man, woman, female, male, boy, girl} address the gender. Notice that a question may fall into more than one category. These categories will be used later to analyze the models.



We did not incorporate any bias towards gender, age, or race during data collection, and tried to be as inclusive as possible in gathering images and formulating questions. We include and balance questions that address different ages and genders. The age groups are \texttt{(baby, 26)}, \texttt{(kid, 42)}, \texttt{(children, 26)}, \texttt{(Teenager, 5)}, \texttt{(Young, 16)}, and \texttt{(old, 12)}. The gender groups are \texttt{(woman, 350)}, \texttt{(women, 38)}, \texttt{(man, 448)}, and \texttt{(men, 79)}. We did not include any question that ask about race.
These issues are more important to address over large training sets. This is because sometimes models trained on such datasets are directly deployed in the real-world.

The BinaryVQA dataset is substantially different from the VQA v2 validation set (the real images) measured in terms of the Fr\'echet Inception Distance (FID)~\cite{heusel2017gans}. The FID is equal to 50.9 indicating a large distribution shift, and hence high diversity (using 7K images). To put this number in perspective, the FID between VQA v2's validation
and its test set is approximately 23.8. Notice that the lower the FID, the more similar the two distributions.



\vspace{-5pt}
\section{Analyses and Results}
\vspace{-5pt}
To see how well the state of the art VQA models perform on our dataset\footnote{We used a 12 GB NVIDIA Tesla K80 GPU to do the experiments.}, we choose the OFA model~\cite{wang2022ofa} which is currently the leading scorer on the VQA v2 test-std dataset\footnote{\scriptsize{\url{https://paperswithcode.com/sota/visual-question-answering-on-vqa-v2-test-std}}}. It achieves 94.66\% accuracy on ``yes/no'' questions. We also include a simple baseline model~\cite{antol2015vqa,zhou2015simple} to see whether transitioning from simple to complicated models in VQA has indeed been meaningful\footnote{\url{https://github.com/iamaaditya/VQA_Demo.git}}. To put the results in perspective, we also ran the Pythia model\footnote{\url{https://github.com/Eurus-Holmes/Pythia-VQA}}. In this section, we focus on explaining the results using the OFA model. Summary results for both models are shown in Table~\ref{tab:perfs}.


\begin{figure*}[t]
\vspace{-10pt}
\centering
  \includegraphics[width=.3\linewidth]{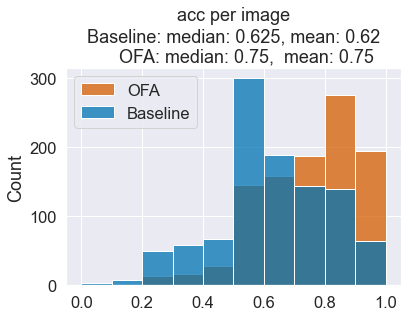} \hspace{10pt}
  \includegraphics[width=.2\linewidth]{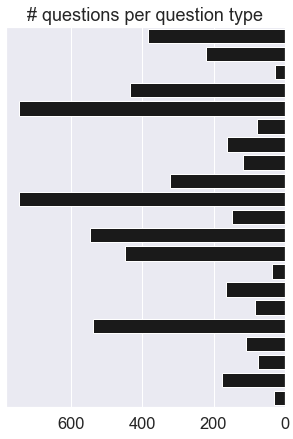}
  \includegraphics[width=.3\linewidth]{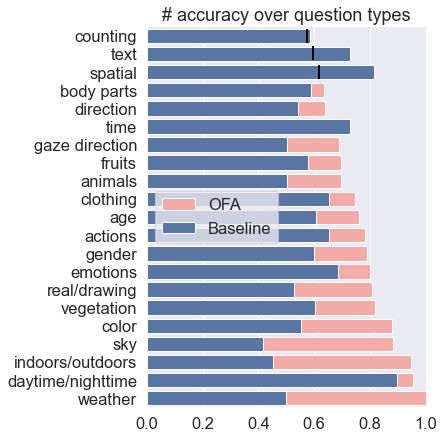}   
\vspace{-7pt}
\caption{\small{Left: Distribution of per image accuracy for models. OFA model is correct $\sim$ 75\% of the time. Middle: Number of questions per question type. Right: Accuracy per question type for models. OFA model does better than the baseline on most of the question types.}}
\label{fig:accs}
\end{figure*}

\begin{figure*}[t]
\vspace{-5pt}
    \centering
    \includegraphics[width=.95\linewidth]{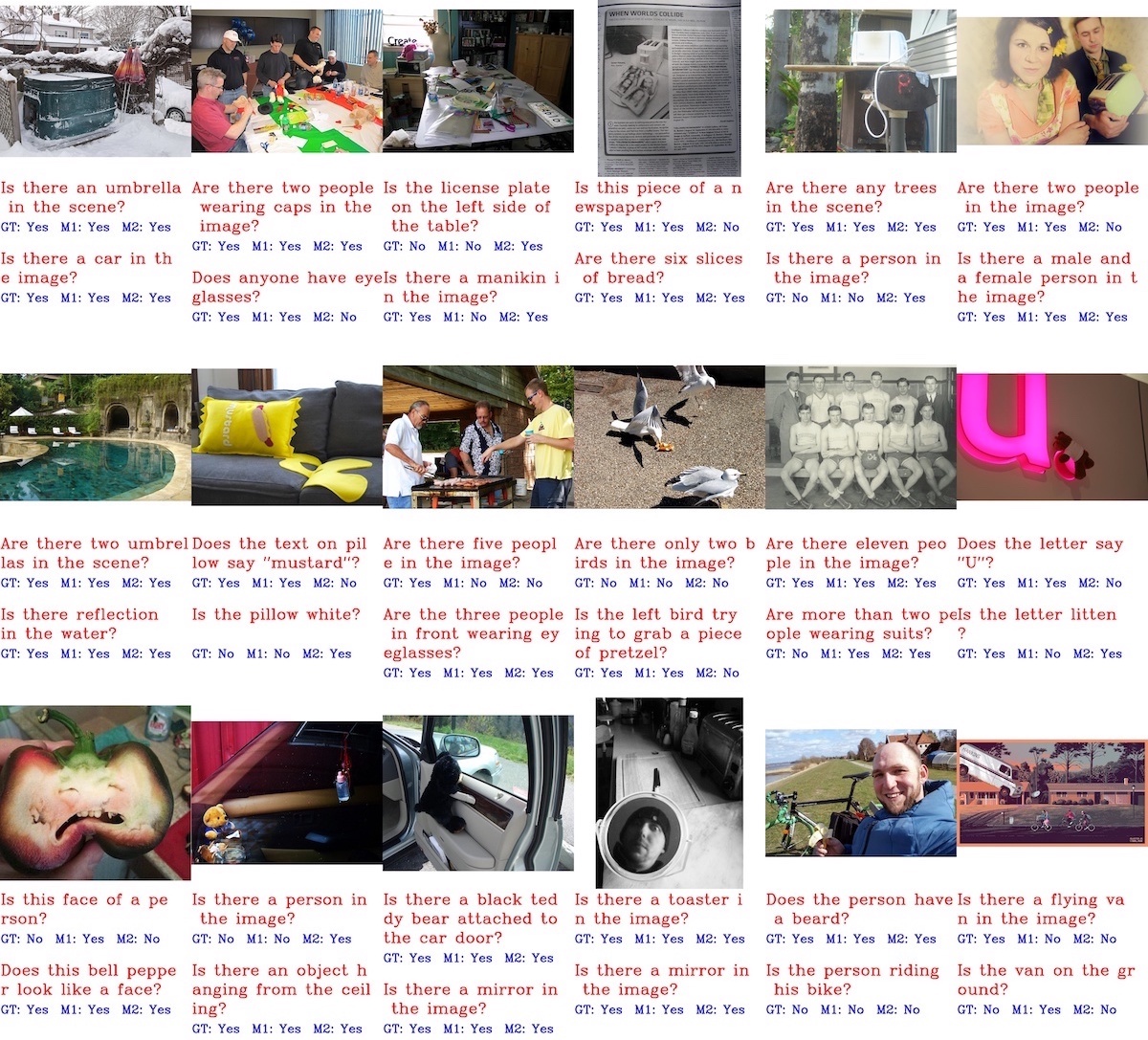}
    \vspace{-5pt}    
    \caption{Sample images along with the question, ground truth answer (GT), prediction of the OFA model (M1) and prediction of the baseline model (M2). See appendix for more examples.}
    \label{fig:samplePredictions}
\vspace{-15pt}
\end{figure*}

The distribution of model scores on the BinaryVQA dataset is shown in the left panel of Fig.~\ref{fig:accs}. The average accuracy of the OFA model is 75\% which is much higher than the 62\% accuracy of the baseline model. The OFA model, however, does significantly worse on our dataset than the VQA v2 dataset (around 20\% absolute performance drop). We attribute this to the more complex nature of the questions and images in our dataset. Sample predictions of both models are shown in Fig.~\ref{fig:samplePredictions}.  


The OFA is able to correctly answer all questions for 160 images (15.6\%) whereas the baseline is right for only 50 images (4.8\%). The OFA model fails all questions over 314 images (30.7\%) while the baseline answers all questions wrong over 673 images (65.7\%).

Performance of the models over question types is shown in the right panel of Fig.~\ref{fig:accs}. The OFA model does better than the baseline in the majority of the question types. It performs below the baseline model over counting (57.2\%), text (59.7\%), and spatial (63\%) categories. It does, however, perform very well on weather (100\%), daytime/nighttime (95.5\%) and indoors/outdoors (96\%) categories. Surprisingly, the OFA model does relatively well in answering questions pertaining to gaze direction (68.7\%) without using any ad-hoc module to process faces, eyes, and gaze angles. The same argument holds over the real/drawing category (80.6\%). We find that models have indeed improved drastically over the years, but there is still a large gap to close. Further, our dataset is significantly harder than the VQA v2 dataset (in ``yes/no'' questions) making it a great auxiliary test set to the existing ones.

\begin{figure}
\centering
\includegraphics[width=.5\linewidth]{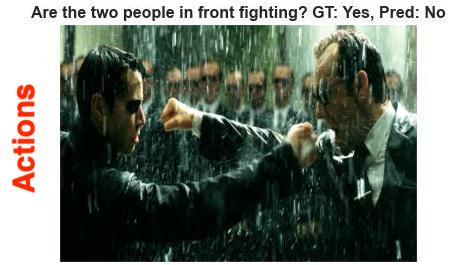} \hspace{-20pt}
\includegraphics[width=.5\linewidth]{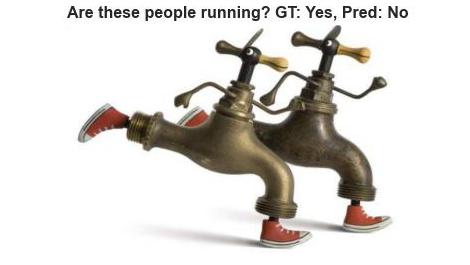}  \\
\includegraphics[width=.5\linewidth]{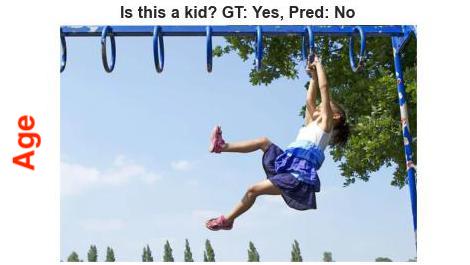} \hspace{-20pt}
\includegraphics[width=.5\linewidth]{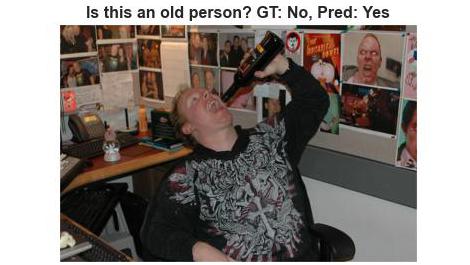} \\
\includegraphics[width=.5\linewidth]{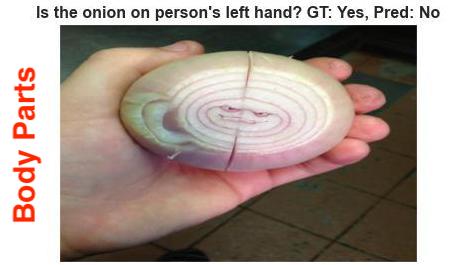} \hspace{-20pt}
\includegraphics[width=.5\linewidth]{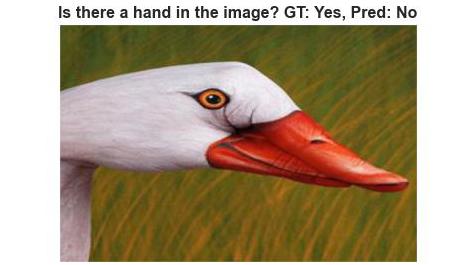}  \\
\includegraphics[width=.5\linewidth]{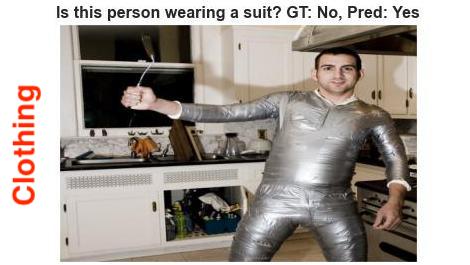} \hspace{-20pt}
\includegraphics[width=.5\linewidth]{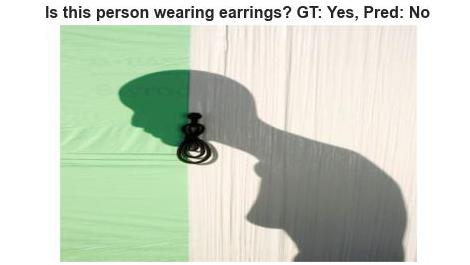}  \\
\includegraphics[width=.5\linewidth]{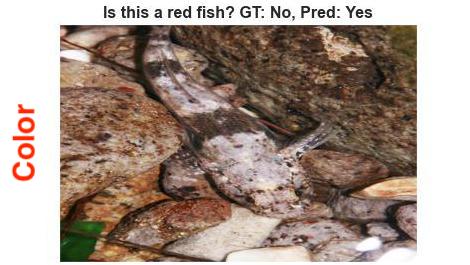} \hspace{-20pt}
\includegraphics[width=.5\linewidth]{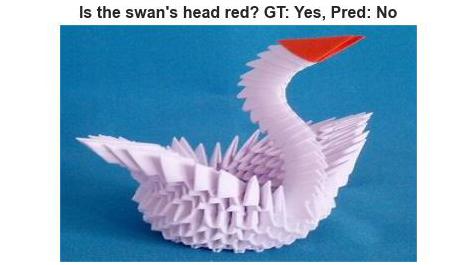} \\
\includegraphics[width=.5\linewidth]{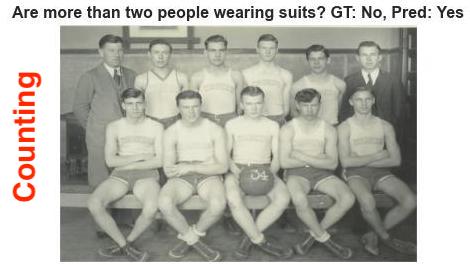} \hspace{-20pt}
\includegraphics[width=.5\linewidth]{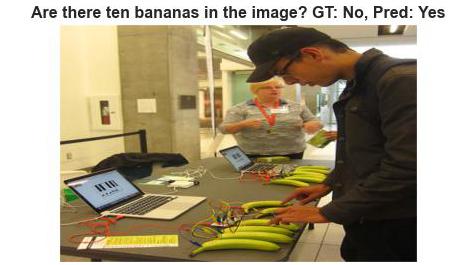}  


\caption{Failure cases of the OFA model over different categories of the BinaryVQA dataset.}
\label{fig:failures}
\vspace{-15pt}
\end{figure}

We found that models perform about the same over real images, paintings, or drawings. OFA model scores $\sim$ 74.12\% over the paintings or drawings (568 questions across 69 drawings/paintings) which is slightly lower than its 75.47\% accuracy on real images (7,232 questions over 955 images). The corresponding numbers for the baseline model are 60.03\% and 63.47\%. The OFA model is correct in answering the counting questions 57.2\% of the time. This model is accurate 69\% of the time over the number category on the VQA v2 dataset. Some difficult questions for the OFA model are shown in Fig.~\ref{fig:failures} over different categories.










\subsection{Model interpretability} 
\vspace{-5pt}
VQA models are very efficient in answering the questions, but how much do they really understand the images? Are their answers grounded on image content, or are merely due to some correlations? Several attempts have been made to address this (\eg~\cite{agrawal2016analyzing,goyal2016towards}) and limiting the image area to a spatial location as is done here (\ie images containing the blue rectangles) is one way to do so. In this section, we propose a new way to interpret the models by masking the image content and study its effect. 
To this end, we run the OpenCV face detector~\cite{viola2001rapid} and mask the faces in images. We then evaluate the OFA model on these images and plot the performance per category as shown in the left panel of Fig.~\ref{fig:extra_res}. Notice that here we limit our analysis to those images for which at least one face is detected (309 out of 1024 images). Some question categories that highly depend on face information such as ``gaze direction'', ``age'', ``gender'', and ``emotions'' are severely degraded, which suggests that models indeed use the right information. Degradation or enhancement over some categories such as ``text'' or ``animals'' may be partially attributed to the false detections of the face detector. This, however, needs further investigation. Note that our masking approach can also be extended to more common objects.


\subsection{Impact of question length on accuracy}
\begin{figure*}[t]
\vspace{-5pt}
\centering
  \includegraphics[width=.95\linewidth]{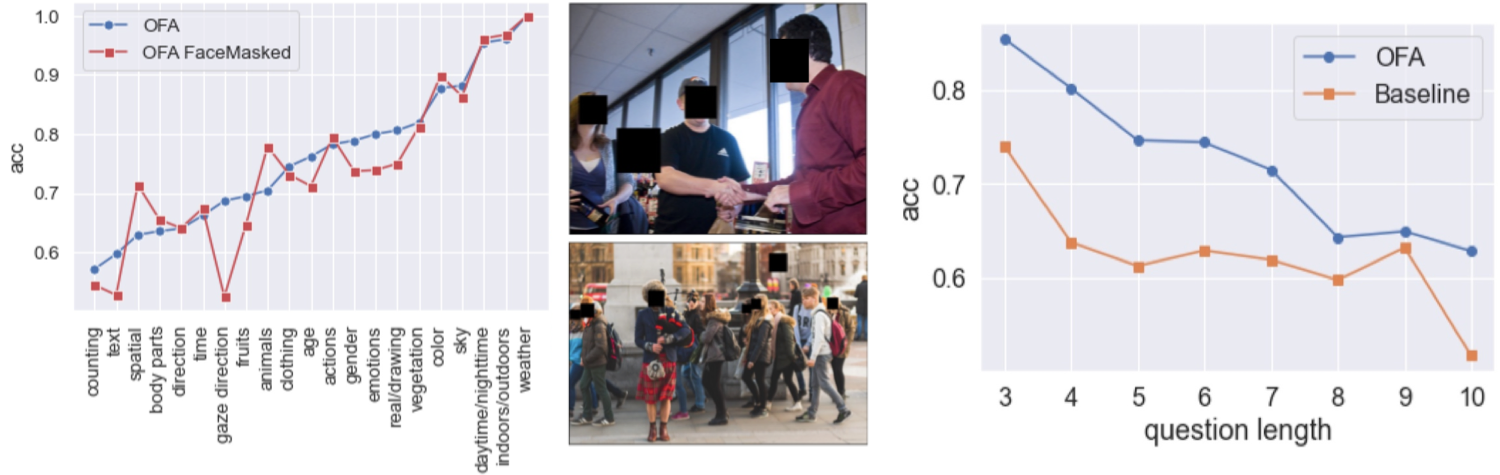} 
\vspace{-5pt}
\caption{Left: Performance of the OFA with and without faces masked. Sample images with faces masked are also shown. Right: Performance of the OFA model as a function of question length.}
\label{fig:extra_res}
\vspace{-10pt}
\end{figure*}

\vspace{-3pt}
Questions in VQA datasets have different levels of complexity. Intuitively, a longer question may be harder to answer than a short one, since it involves unpacking and understanding the dependencies among words in the sentences and their corresponding objects in the image. The right panel of Fig.~\ref{fig:extra_res} shows the model accuracy as a function of question length. Due to rarity, questions longer than 10 words are discarded (only 150 occurrences). As it can be noticed, accuracy decays as the question length grows. The mean accuracy of the OFA model over questions less than 8 words is 72.3\%. Its accuracy over questions longer than 8 words (and less than 10) is 51.6\%. The corresponding numbers for the baseline model in order are 62.3\% and 52.8\%. This result corroborates previous findings over the VQA dataset and shows that models underperform over longer questions. Since our dataset contains longer questions than the VQA dataset, it can better test this aspect of models.






\subsection{Analysis of ``yes'' bias in models} 
\vspace{-5pt}
VQA datasets usually contain more questions with ``yes'' answers than questions with ``no''  answers. This is partially due to the tendency of annotators to query the existing content in images. Consequently, a smart chance model that often produces positive answers may win over a sophisticated model. One approach to combat this issue, as is done over the VQA v2 dataset, is to balance the distribution of positive and negative questions. Here, we introduce a new score called ``ShuffleAcc'' to automatically address this. 
A subset of $2n$ questions consisting of $n$ positive and $n$ negative questions are randomly selected (here $n=2000$). The average model accuracy over $m$ such subsets is then computed (here $m=50$). A model that consistently generates a ``yes'' (or ``no'') answer will achieve 50\% accuracy. The same argument holds for a model that randomly chooses ``yes'' 50\% of the time.
The ShuffleAcc scores of OFA and baselines models in order are 75\% and 62.4\% which are about the same as their performance using the traditional accuracy score. This entails that these models do not suffer from inherent biases towards positive answers.

\subsection{Sensitivity to spelling and grammar errors}
\vspace{-5pt}
Studies on understanding and evaluating VQA models have been primarily focused on the visual component of VQA. Less attention, however, has been paid to diagnosing errors in the NLP component, in particular the sensitivity of models to perturbations on asked questions. This is particularly important to study since we know humans are still able to correctly answer questions even in presence of significant spelling and grammar mistakes, so long the meaning of the question remains the same. Here, we study three simple perturbations that are unlikely to change the answer.



\textbf{Within-word character swap.} Here, we first randomly select a word (with length $>3$) in the question. Next, we randomly choose two characters in this word and swap them. For example, the question ``Is there a person in the image?'' will turn into ``Is there a peosrn in the image?''. We then evaluate the OFA model by varying the number of words, from 1 to 3, for which we swap two characters. 
OFA accuracy drops to 61.4\% with swap in one word, 53.5\% with swaps in two words, and 49.1\% with swaps in three words. These results clearly show that spelling errors drastically hinder the models. Humans often do not notice these changes during reading.

To test whether this result generalizes to other datasets, we repeated these experiments over the VQA-v2 test set. The accuracy of the OFA model drops to 91.7\%. This number drops to 84.7\% with swap in one word, 77.3\% with swaps in two words, and 65.5\% with swaps in three words. Similar observations are made for the baseline model.

\textbf{Omission of the articles.}
Here, all the articles (``the'', ``a'', ``an'') are removed from the question. For instance, the question ``Is the person on the right holding a camera?'' will be converted to ``Is person on right holding camera?''. The performance of the OFA model drops to 73.8\% indicating that this model, similar to humans, is robust to the omission of the articles. 


\textbf{Negating the question.} Questions in the BinaryVQA dataset are formulated positively without using the word ``not''. Logically, if the question is negated the answer should also be negated\footnote{Of course there are exceptions in the conversational language, \eg Isn't there a person in the room? Answer: No! (assuming there are no people in the room).} For example, if the answer to the question ``Is there a firefighter on the crane?'' is ``yes'', then the answer to the question ``Is there not a firefighter on the crane?'' should be ``no''. For this analysis, we focus only on ``Is there'' type questions. Out of 1,841 such questions, the OFA model maintained its decision in 738 cases when the question was negated. This amounts to about 40\% of the cases, which is far above 0\%. Ideally, the model should always reverse its decision.


\subsection{Ablation analyses} 
\vspace{-3pt}
Following our interpretability analysis above, here we conduct two analyses which can be considered as sanity checks or baselines for models. Models can be right for wrong reasons, and vice versa. In the first analysis, we ask all the questions over a black image or a white noise image. The OFA model performs well below chance, about 36.4\% and 36.89\% over these images, respectively. This indicates that this model indeed requires the image to produce the right answer.



The second analysis investigates whether a model can consistently produce the ``no'' answer to questions for which we know the answer is surely ``no''. We asked 15 questions in the form of ``Is there a/an X in the image?'' where X represents one of the following objects \texttt{`white orange', `dragon', `blue horse', `backgammon board', `parrot', `boxer dog', `ostrich', `dinosaur egg', `galaxy', `mermaid', `telescope', `unicorn', `centipede', `yellow cow', `yeti'} over all the 1024 images. The mean accuracy of OFA model across all 15 $\times$ 1024 questions is 93.1\% using original images. The breakdown per each of these questions is shown in Appendix C. Interestingly, when we asked these questions on white noise images, the accuracy jumped to 100\%. These results again demonstrate that the OFA model indeed highly relies on the image content.

\begin{table}[t]
\begin{center}
\begin{scriptsize}
\renewcommand{\tabcolsep}{2pt}
\renewcommand{\arraystretch}{.8}
\begin{tabular}{l|c|c|c| c | c || c}
Model & Avg Acc.  & ShuffledAcc & Char Swap & Article & Question$^{*}$  & Acc on \\
& &  & (one word) &  Omission & Negation (\%) & VQA v2$^{+}$ \\
\hline \hline

Baseline & 62.5  & 62.4  &  51.5 & 59.3 & 35 & 80.5\\
OFA & 75  & 75  & 61.4 & 73.8 & 40 & 94.66\\
Pythia & 72.1  & 72.2  & 58.8 & 69.4 & 46 & 86.7$^{\dagger}$\\

\end{tabular}
\end{scriptsize}
\end{center}
\vspace{-10pt}
\caption{\scriptsize{Summary of model performance on BinaryVQA dataset. \\
$*$ = Percentage of questions for which the model retained its answer after negation. \\
$+$ = Human perf. is about 95.48 from \url{https://visualqa.org/roe.html} \\ 
$\dagger$ = Pythia v0.1 the winning entry in 2018 VQA benchmark \url{https://visualqa.org/roe_2018.html} \\
}}
\vspace{-20pt}
\label{tab:perfs}
\end{table}

\section{Discussion and Conclusion}


Understanding complex questions in VQA is a big challenge, so is the understanding of complex scenes. Our dataset is better suited to address the latter, whereas other datasets can address the former. It can be used to test models that already perform above 95\% on binary questions of VQA-v2 dataset. Our dataset contains a lot of questions which are really challenging and need close examination of the image to be answered. Such questions ask about non-standard objects, surreal imagery, and/or other oddities (\eg an eagle with a banana for a beak, water spout wearing sneakers, an odd clothespin-like object on one side and spoon on the other, a face with multiple pairs of eyes).

We share a zip file containing images, questions, metadata, and detailed documentation. BinaryVQA is licensed under Creative Commons Attribution 4.0 (Appendix E).

\medskip

{\small
\bibliographystyle{ieee_fullname}
\bibliography{egbib}
}






\newpage
\appendix


\newpage
\appendix

\section{Samples images, questions, and answers from the BinaryVQA dataset}

\begin{figure*}[htbp]
    \centering
        \vspace{-15pt}    
    \includegraphics[width=.9\linewidth]{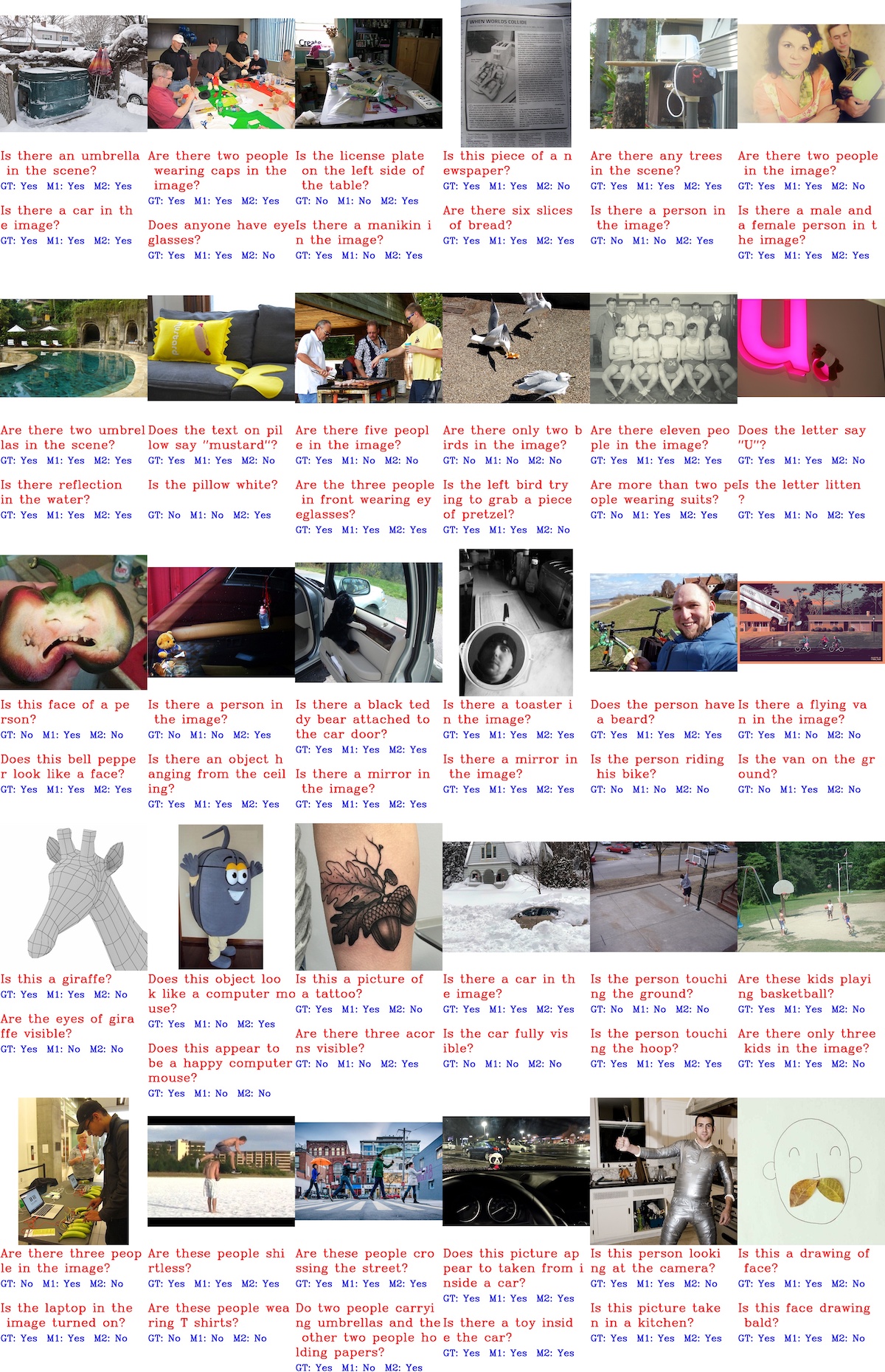}
    \caption{Sample images along with the question, ground truth answer (GT), prediction of the OFA model (M1) and prediction of the baseline model (M2).}
    \vspace{-35pt}    
\end{figure*}

\begin{figure*}[htbp]
    \centering
        \vspace{-15pt}    
    \includegraphics[width=.9\linewidth]{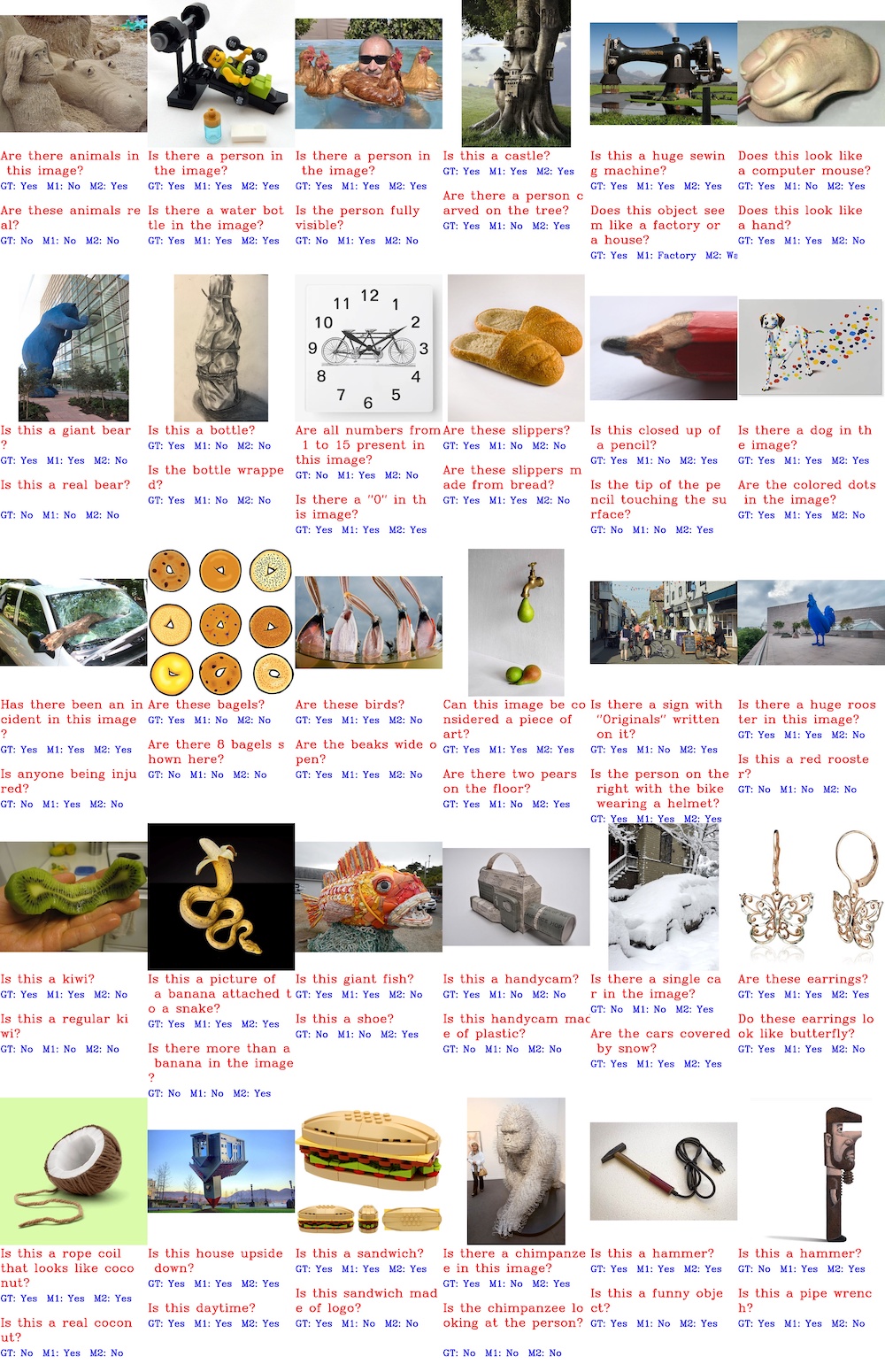}
    \caption{Sample images along with the question, ground truth answer (GT), prediction of the OFA model (M1) and prediction of the baseline model (M2).}
    \vspace{-35pt}    
\end{figure*}

\begin{figure*}[htbp]
    \centering
    \includegraphics[width=.9\linewidth]{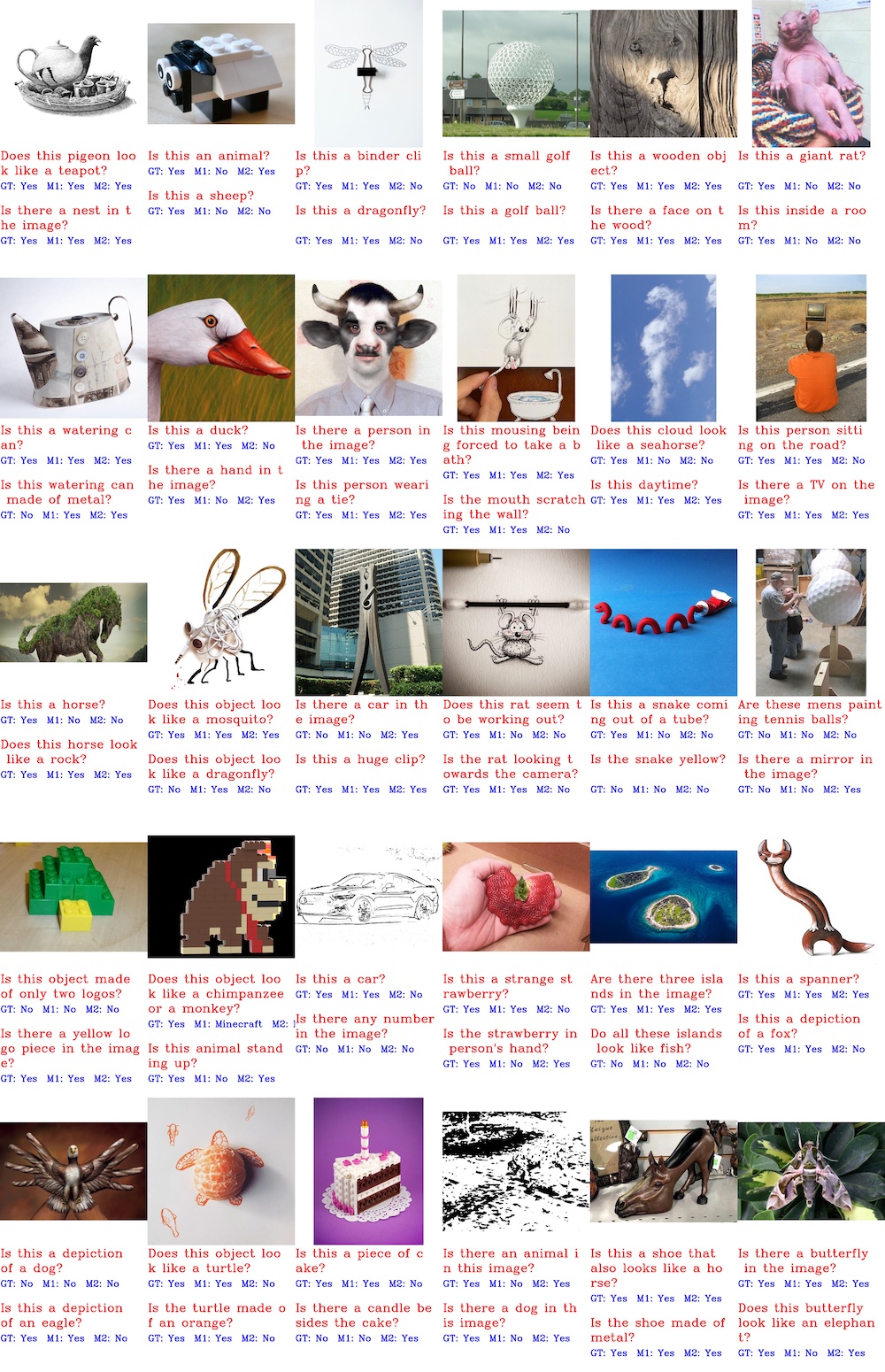}
    \vspace{-15pt}    
    \caption{Sample images along with the question, ground truth answer (GT), prediction of the OFA model (M1) and prediction of the baseline model (M2).}
\end{figure*}

\begin{figure*}[htbp]
    \centering
    \includegraphics[width=.9\linewidth]{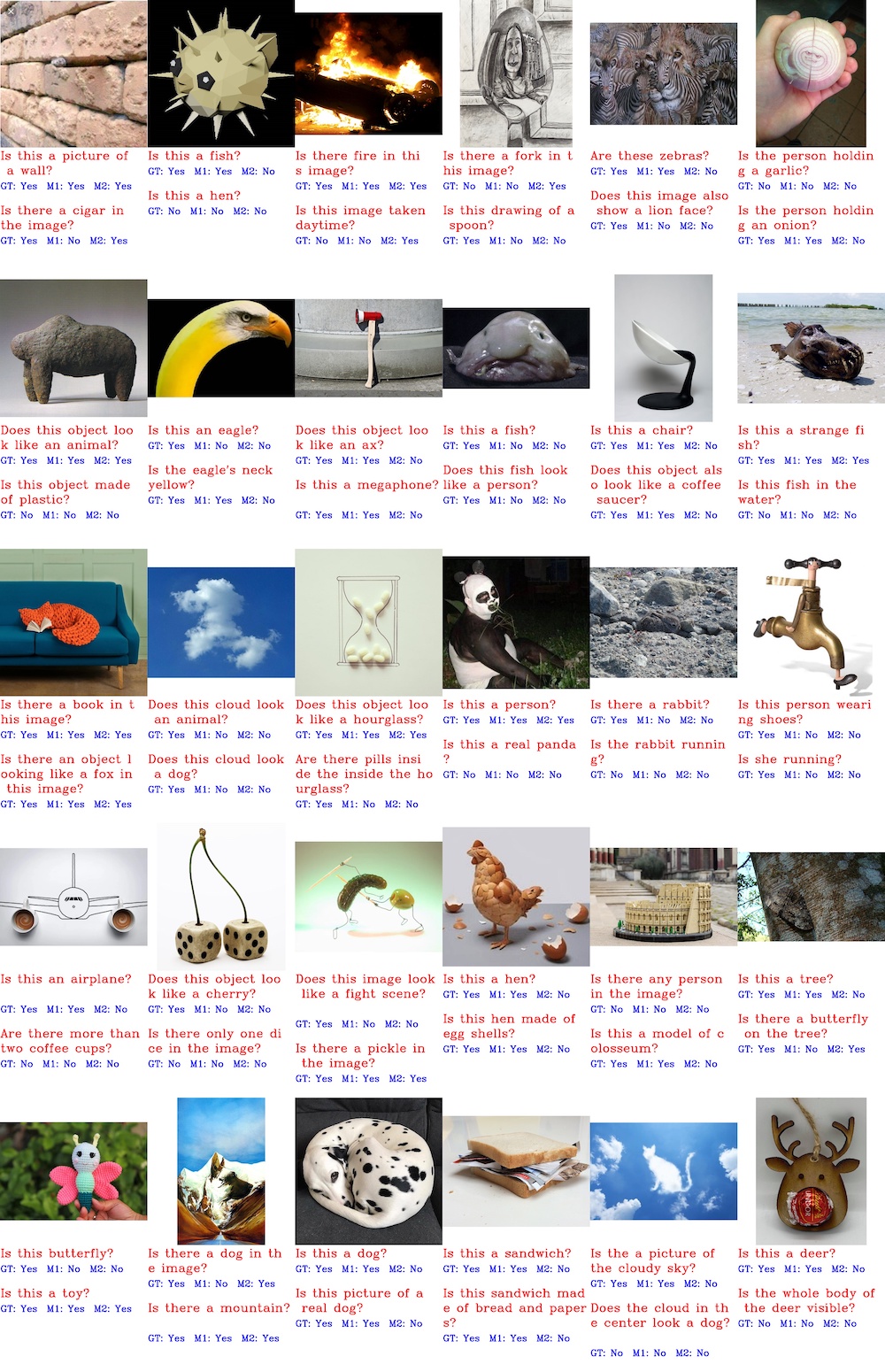}
    \vspace{-15pt}    
    \caption{Sample images along with the question, ground truth answer (GT), prediction of the OFA model (M1) and prediction of the baseline model (M2).}
\end{figure*}

\begin{figure*}[htbp]
    \centering
    \includegraphics[width=.9\linewidth]{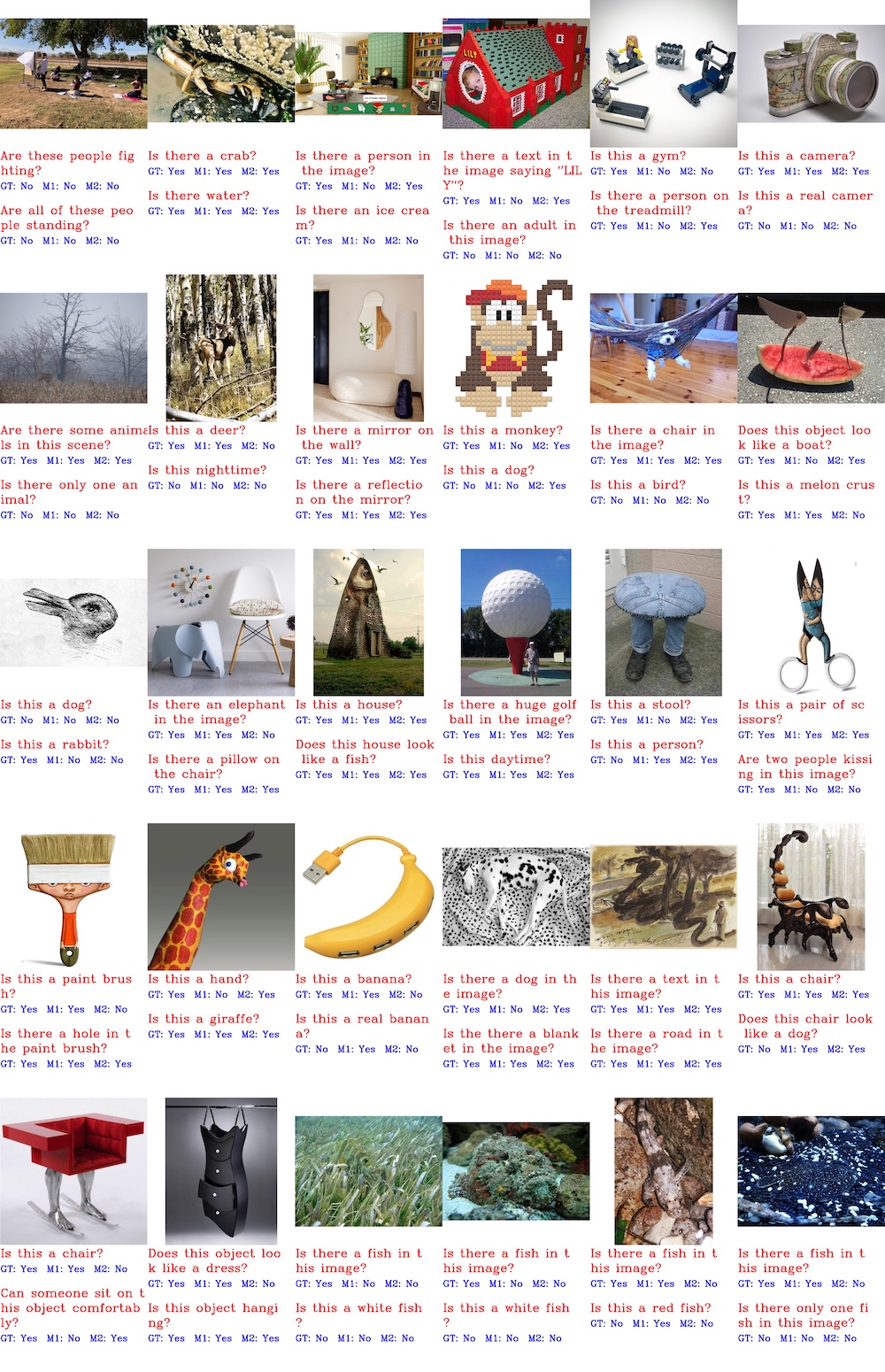}
    \vspace{-15pt}    
    \caption{Sample images along with the question, ground truth answer (GT), prediction of the OFA model (M1) and prediction of the baseline model (M2).}
\end{figure*}

\begin{figure*}[htbp]
    \centering
    \includegraphics[width=.9\linewidth]{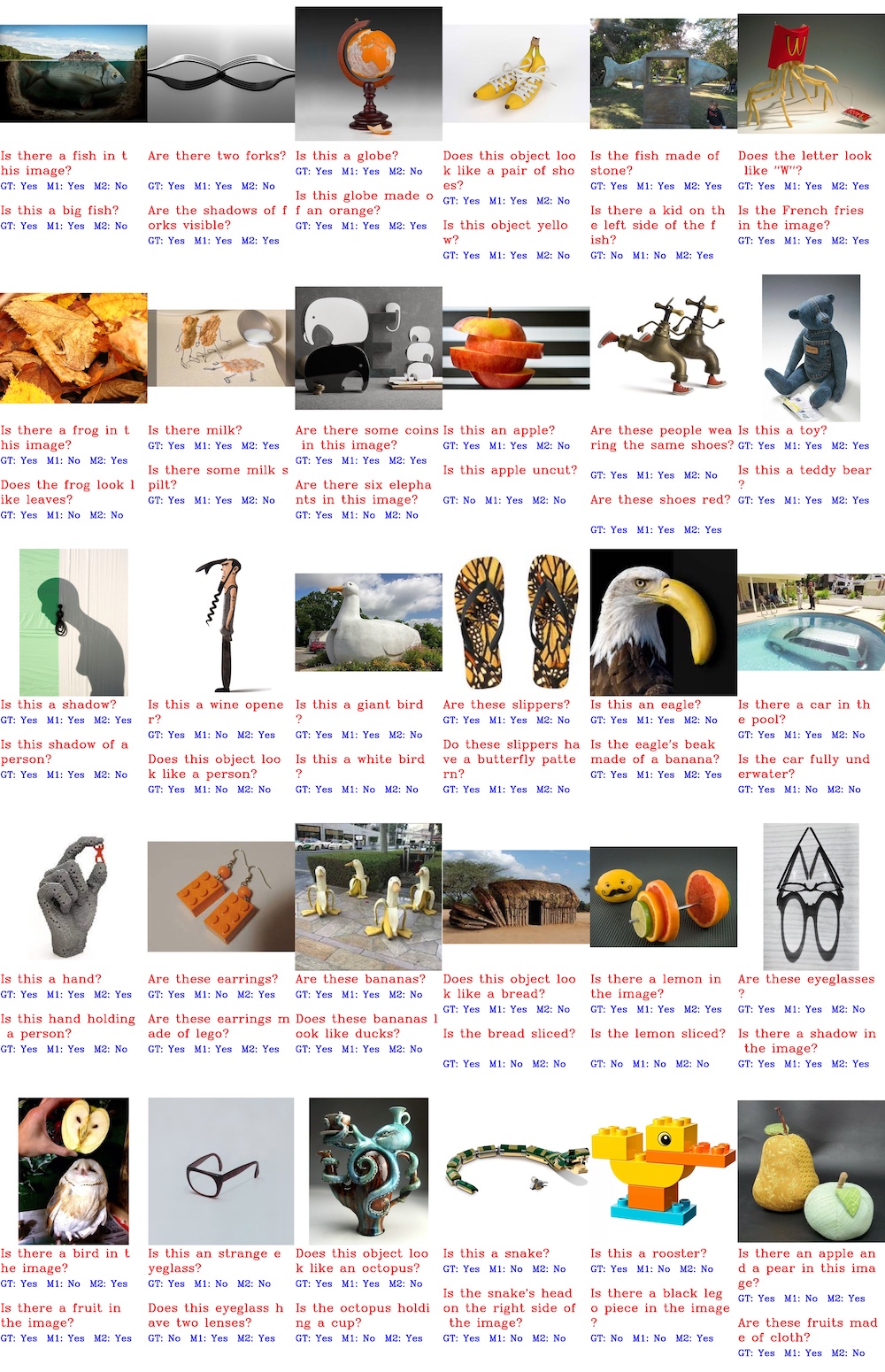}
    \vspace{-15pt}    
    \caption{Sample images along with the question, ground truth answer (GT), prediction of the OFA model (M1) and prediction of the baseline model (M2).}
\end{figure*}

\clearpage
\section{Samples images from the BinaryVQA dataset}

\begin{figure*}[htbp]
    \centering
    \includegraphics[width=\linewidth]{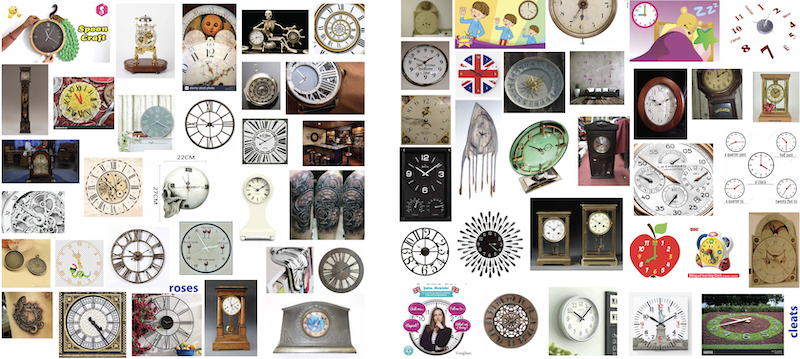}
    \caption{Sample clock images with Roman numerals (left) and English numerals (right).}
\end{figure*}

\begin{figure*}[htbp]
    \centering
    \includegraphics[width=\linewidth]{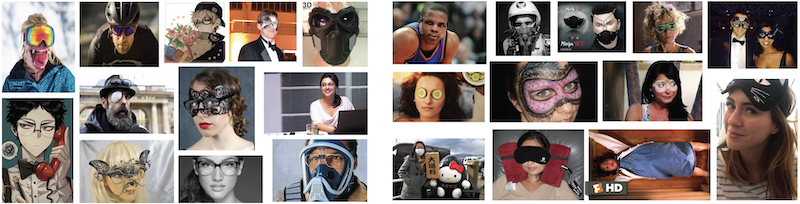}
    \caption{Sample clock images with (left) and without eye glasses (right).}
\end{figure*}

\begin{figure*}[htbp]
    \centering
    \includegraphics[width=\linewidth]{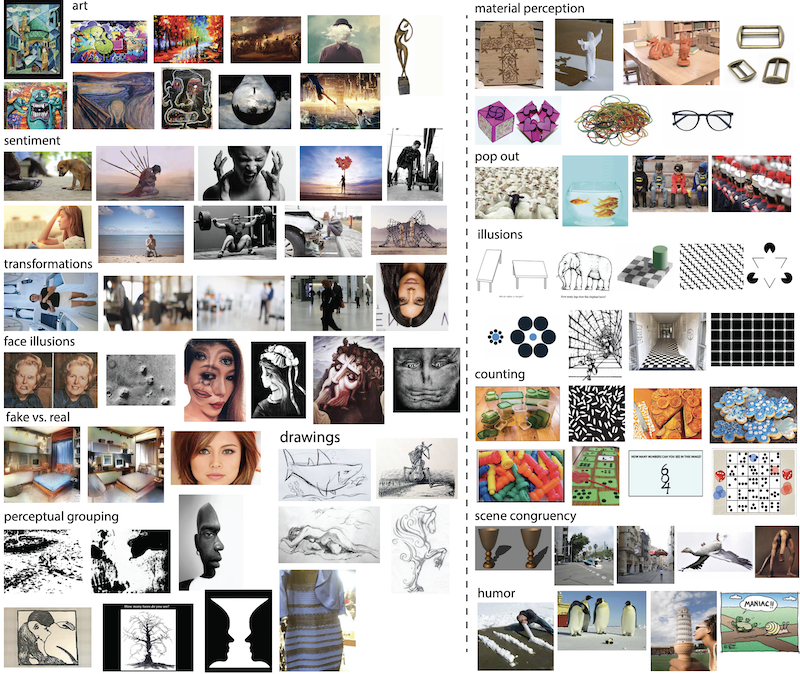}
    \caption{Additional images from the BinaryVQA dataset.}
\end{figure*}

\clearpage

\section{Breakdown over the words in ablation study}
The accuracy of the OFA model over questions asking about existence of non-existing objects in the image. 

\begin{table}[htbp]
\begin{center}
\begin{small}
\begin{tabular}{|l|c|c|}
\hline 
Object & Accuracy over the original image & Accuracy over the white noise image\\
\hline 

white orange & 0.702 & 1\\
dragon & 0.916& 1 \\
blue horse & 0.958& 1 \\
backgammon board & 0.953 & 1\\
parrot & 0.983 & 1\\
boxer dog & 0.965& 1 \\
ostrich & 0.990 & 1\\
dinosaur egg & 0.985& 1 \\
galaxy & 0.863 & 1\\
mermaid & 0.956 & 1\\
telescope & 0.900& 1 \\
unicorn & 0.983 & 1\\
centipede & 0.981& 1 \\
yellow cow & 0.933 & 1\\
yeti & 0.891 & 1\\

\hline 

\end{tabular}
\end{small}
\end{center}
\caption{performance of the OFA model over questions of the type ``Is there a/an X in the image? Replace X with the object name in the first column.}
\end{table}

\clearpage


\section{Data collection}
We adopt the following high-level process to collect the images and (question,answer) pairs. First, we generated some phrases and then searched Flickr or Google search to find matching images. We limited the search results to only those images that had the creative commons licences. Some sample search queries include: ``A couple of kids watching TV in a room while sitting on the floor?", ``A woman looking at the camera while eating a burger?", ``A couple of people in a meeting room?", ``Two people fighting'', ``A cat in the clouds'', ``A sheep made of lego'', ``A man with blond hair'', etc. We then formulated some questions on these images along with answers. The (question,answer) pairs were presented to three AMT workers for further verification. Few questions for which AMT workers did not agree were then corrected. 


Our AMT interface for collecting the verification of our answers to the questions.
Workers were paid 25 cents per question. The experiment took 30 hours per participant.  

\begin{figure}[htbp]
    \centering
    \includegraphics[width=\linewidth]{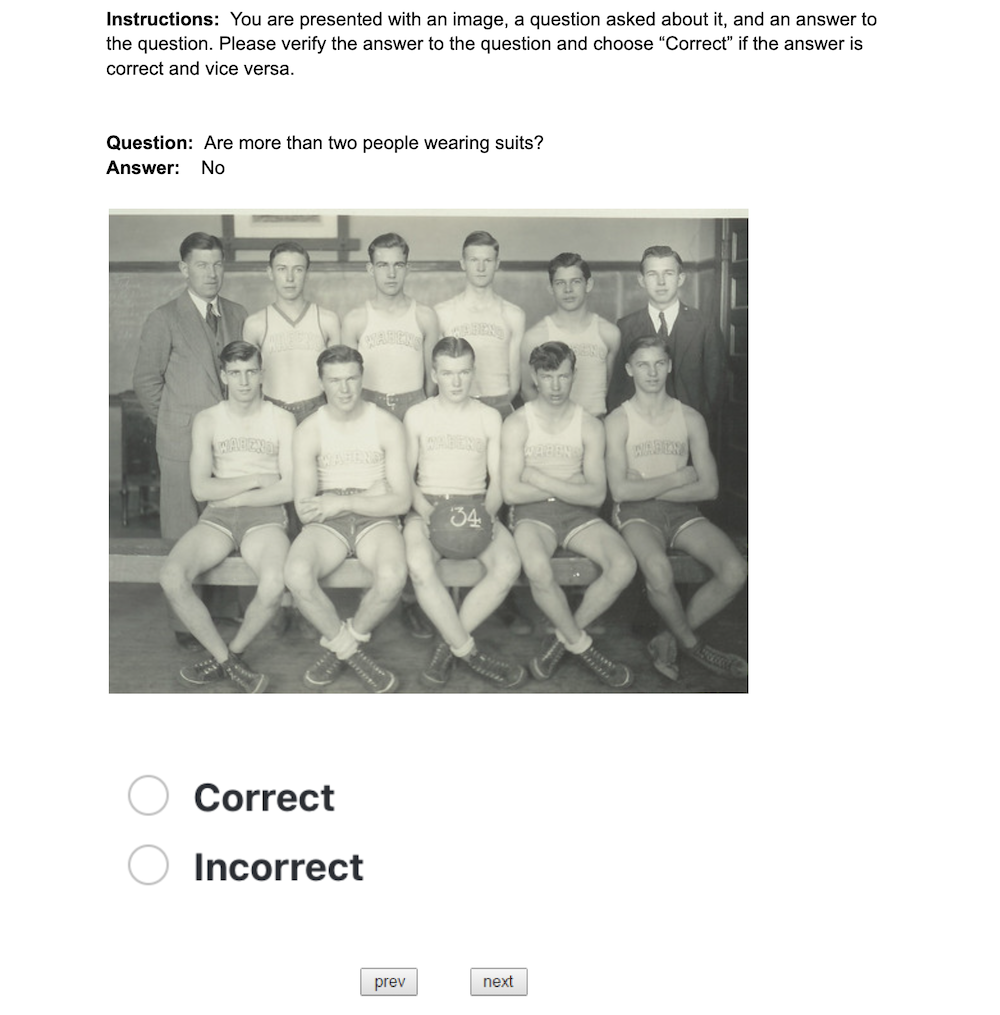}
    \caption{Our AMT interface for collecting the verification of our answers to the questions.}
\end{figure}

We have 17 images (from 0700.jpeg to 0716.jpeg) that have blue rectangles. 25 questions were asked on these rectangles. These questions either asked about an object or a person inside the rectangle (\eg Is there a spatula inside the blue rectangle?) or something about the rectangle itself (Is the blue rectangle on the bottom right corner of the image?).


\clearpage

\section{Dataset License}

BinaryVQA dataset is free to use only for research and academic purposes (not commercial). It is licensed under Creative Commons Attribution 4.0 with three additional clauses:

\begin{enumerate}
    \item BinaryVQA may never be used to tune the parameters of any model.
    \item The images containing people should not to be posted anywhere unless the people in the images are appropriately de-identified. Even in this case, written agreement from dataset creators is required. This is to check whether all the clauses are properly followed.
    
\end{enumerate}

To stop or limit the misuse of our BinaryVQA by bad actors, we have made a dataset request form\footnote{\url{https://bit.ly/3bDY0MS}}. We review the requests that we receive and allow
access for a legitimate use. The dataset we share contains images and questions is a zip file. The package also contains the detailed documentation with all relevant metadata specified to users.

\clearpage
\section{Experimental details and evaluation setup}

We have used the validation set of the balanced real scens from the VQAv2 dataset from \url{https://visualqa.org/download.html}. We are only using the binary questions. Images are resized and normalized. A questionmark is added to the questions if it is missing. BOS and EOS tokens are also added to the question. Model parameters for each of the tested models are listed below.

Parameter settings for VQA baseline: 
\begin{itemize}
    \item VGG\_16 model
    \item 4096 D feature vector for the image representation
    \item Image size 224 x 224
    \item Each word in the question is a Glove vector 300D
\end{itemize}

OFA model:
\begin{itemize}
\item Checkpoint: ofa\_large\_384.pt
\item Images are resized to and normalized
\item A questionmark is added if missing 
\item BOS and EOS tokens are added
\end{itemize}

Pythia model:
\begin{itemize}
  \item TARGET\_IMAGE\_SIZE = [448, 448]
  \item CHANNEL\_MEAN = [0.485, 0.456, 0.406]
  \item CHANNEL\_STD = [0.229, 0.224, 0.225]
\end{itemize}

\end{document}